IAC-23-A3-2B

# Modularity for lunar exploration: European Moon Rover System Pre-Phase A Design and Field Test Campaign Results

**C. Luna*[a], J. Barrientos-Díez [a], M. Esquer [a], A. Guerra [a], M. López-Seoane [a], I. Colmenarejo [a], F. Gandía [a], S. Kay [b], A. Cameron [b], C. Camañes[c], I. Sard[c], D. Juárez[c], A. Orlandi[d], F. Angeletti[d], V. Papantoniou[e], A. Papantoniou[e], S. Makris[e], B. Rebele[f], A. Wedler[f], J. Reynolds[g], M. Landgraf[g]**

[a] *GMV Aerospace and Defence SAU, Calle de Isaac Newton 11, Tres Cantos, Madrid, Spain,* cluna@gmv.com.
[b] *GMV NSL Ltd, Airspeed 2, Eighth Street, Harwell Campus, Oxfordshire, UK, OX11 0RL*
[c] *AVS Added Value Solutions, Elgoibar, Spain*
[d] *OHB System AG, Manfred-Fuchs-Strasse 1, 82234 Wessling, Germany.*
[e] *Hellenic Technology of Robotics, Kfisias Ave 188, Athens 14562, Greece.*
[f] *German Aerospace Center (DLR), Institute of Robotics and Mechatronics, Münchener Str. 20, 82234 Weßling, Germany.*
[g] *ESTEC, ESA, Keplerlaan 1, 2201 AZ Noordwijk, The Netherlands.*
* Corresponding Author

**Abstract**

The European Moon Rover System (EMRS) Pre-Phase A activity is part of the European Exploration Envelope Programme (E3P) that seeks to develop a versatile surface mobility solution for future lunar missions. These missions include: the Polar Explorer (PE), In-Situ Resource Utilization (ISRU), and Astrophysics Lunar Observatory (ALO) and Lunar Geological Exploration Mission (LGEM). Therefore, designing a multipurpose rover that can serve these missions is crucial. The rover needs to be compatible with three different mission scenarios, each with an independent payload, making flexibility the key driver. This study focuses on modularity in the rover's locomotion solution and autonomous on-board system. Moreover, the proposed EMRS solution has been tested at an analogue facility to prove the modular mobility concept. The tests involved the rover's mobility in a lunar soil simulant testbed and different locomotion modes in a rocky and uneven terrain, as well as robustness against obstacles and excavation of lunar regolith. As a result, the EMRS project has developed a multipurpose modular rover concept, with power, thermal control, insulation, and dust protection systems designed for further phases. This paper highlights the potential of the EMRS system for lunar exploration and the importance of modularity in rover design.

## 1. Introduction

The European Moon Rover System (EMRS) Pre-Phase A activity fits in the framework of the European Exploration Envelope Programme (E3P), which is currently calling for a versatile surface mobility solution to further advance lunar exploration activities. Future lunar missions are indeed envisioned to rely on a multi-mission landing capability, namely the Argonaut, formerly the European Large Logistic Lander (EL3), to carry out different science applications. Among them, three missions – currently in pre-Phase A status - require a mobile solution:

- Polar Explorer (PE),
- In-Situ Resource Utilization (ISRU)
- Astrophysics Lunar Observatory (ALO)
- Lunar Geological Exploration Mission (LGEM)

In this scenario, the main objective of this study is to preliminary design the EMRS to achieve modularity and flexibility[1] of use in different mission configurations, while obtaining a good balance between mission versatility and system optimality.

### 1.1 Polar Explore (PE)

The Polar Explorer mission represents a pioneering endeavour centred on the exploration and analysis of water ice deposits and volatile compounds. This ambitious mission is further characterized by its mobility, which is integral to its core concept, drawing inspiration from the rich legacy of prior polar exploration missions. Moreover, the mission encompasses a multifaceted approach, aligning with various objectives within the broader scope of European space exploration.

As part of its scientific arsenal, the Polar Explorer mission incorporates an array of complementary science packages designed to operate both within the confines of the lander and on the lunar surface itself. These meticulously designed packages, in conjunction with the mobile prospecting element, collectively constitute a formidable scientific apparatus poised to tackle pivotal questions within the realms of lunar science and exploration. These supplementary science packages include Geophysics Station, Lunar Surface Environment Package, Exobiology Exposure Package. Radio Receiver Antennae

Each of these science packages is strategically integrated into the Polar Explorer mission to synergistically complement the mobile prospecting element's capabilities. Together, they form an interdisciplinary scientific consortium poised to address fundamental questions and contribute to the advancement of lunar exploration and our understanding of the moon's geophysical and astrobiological dimensions.

### 1.2 Astrophysical Lunar Observatory (ALO)

The Astrophysics Lunar Observatory (ALO) embodies a pioneering mission concept with the primary aim of elucidating the enigmatic aspects of the deep and ancient universe. This ambitious effort harnesses the potential of the electromagnetic spectrum, specifically focusing on the realm of radio waves. At its core, the overarching objective of the ALO mission is to establish a cutting-edge, low-frequency radio interferometric array on the distant lunar terrain.

This strategic placement on the far side of the Moon offers a distinct advantage, capitalizing on the unparalleled isolation that this lunar vantage point provides. It acts as an impervious shield, safeguarding the observatory from the detrimental effects of terrestrial radio frequency interference, as well as the disruptive influences of auroral kilometric radiation and plasma noise originating from the solar wind. Consequently, ALO is primed to undertake a series of observations that are characterized by the remarkable capability to circumvent the constraints of sky noise, particularly in the sub-MHz frequency range.

### 1.3 In-Situ Resources Utilisation (ISRU)

The In-Situ Resource Utilization (ISRU) pilot plant mission is strategically designed to harness lunar resources, thereby laying the groundwork for future manned lunar missions. By extracting essential resources directly from the lunar soil, this mission not only enhances the sustainability of human lunar exploration but also significantly mitigates the financial burden associated with transporting resources from Earth. Furthermore, the acquisition of in-situ resources holds the potential to extend mission durations and forms the cornerstone for the establishment of a potentially permanent and economically viable human outpost on the moon.

The primary utilization scenarios under evaluation encompass the production of propellant (as the paramount objective), resupplying life support essentials (as a secondary goal) and generating fuel cell reactants (tertiary focus). These critical resources are pivotal to sustaining lunar missions and reducing dependence on Earth-based supply chains.

The mission's targeted area is the Schrödinger Crater, carefully chosen for its scientific significance and resource potential.

The mission comprises two essential components: a stationary pilot plant and a mobile rover equipped with excavation tools. The rover's primary role is to excavate lunar regolith, transporting it to the pilot plant for resource extraction. Additionally, the pilot plant functions as a hub for the rover, providing recharging capabilities, a vital requirement due to the unreliable nature of conventional rover-mounted power sources like solar panels, which are prone to dust exposure resulting from excavation activities.

The primary objective of the rover is to collect up to 300 kg of lunar regolith per excavation cycle and deliver it to the pilot plant. Over the course of its operational lifespan of one full year, the rover will execute these delivery cycles at regular intervals, ensuring the sustained operation of the pilot plant. This cumulative effort will result in 42 cycles, translating to a total regolith delivery quantity of 12,600 kg. Additionally, the rover is tasked with the removal of processed regolith ejected by the pilot plant, maintaining operational efficiency and cleanliness.

### 1.4 Lunar Geological Exploration Mission (LGEM)

The Lunar Geology Exploration Mission (LunarGEM) is a mission concept within the Terrae Novae program, designed to prepare for future lunar science exploration. LunarGEM offers adaptability, allowing for alternative delivery methods and hybrid mission configurations. It is projected to be ready for the Argonaut-2 mission or later, potentially by 2032.

The primary objective of LunarGEM is to lay the groundwork for future lunar science exploration. This includes informing ongoing EMRS and EL3 industrial studies and shaping decisions related to lunar exploration preparation activities in subsequent E3P programmatic phases. The mission's key feature is the inclusion of a LIDAR-based instrument onboard the rover, facilitating the creation of high-resolution terrain models through laser-based distance measurements.

Operational constraints dictate that the rover functions during periods of illumination when there is sufficient sunlight for power generation and allows communication with Earth using Direct To Earth (DTE) technology. The rover is equipped to handle relatively steep slopes, with maximum acceptable pitch and roll angles of 15 degrees, and it can achieve a minimum traverse speed of 2.5 cm/s on plain terrain. Notably, LunarGEM can provide highly detailed lunar surface images, with a spatial resolution of at least 2 meters per pixel.

## 2. State-of-the-art

In the realm of robotic mobility systems for Mars and Moon exploration, several cutting-edge rover developments are shaping the current state of the art. These rovers, either slated for deployment in the near future or already in operation, are at the forefront of planetary exploration.

VIPER (NASA): Currently under development by NASA, the VIPER rover is primed for a lunar mission aimed at exploring the Moon's southern pole in pursuit of frozen water deposits. Anticipated for a late 2024 lunar landing [2], VIPER boasts a weight of 430 kg, dimensions measuring 1.53m x 1.53m in footprint, and a height of 2.45m [2]. Key scientific instruments include a one-meter-long regolith and ice drill, a Neutron Spectrometer System, a Near-Infrared Volatiles Spectrometer, and a Mass Spectrometer [3]. Its locomotion system features four identical modules, each equipped with wheels, independent steering capabilities, and an active suspension system. These modules are tailored for specific locomotion modes, allowing VIPER to traverse low compaction sand effectively, with an average maximum speed of 0.2 m/s [3].

PERSEVERANCE (NASA): Currently operational on Mars, the Perseverance rover conducts a multifaceted scientific investigation, including the search for signs of past microbial life, the collection of Mars regolith samples, and groundbreaking experiments such as the helicopter Ingenuity's historic flight [4][5]. This rover boasts dimensions of 2.9 meters in length, 2.7 meters in width, 2.2 meters in height, and a weight of 1025 kg [6]. Equipped with a robotic arm housing an interchangeable drill system, along with various spectrometers, cameras, environmental sensors, and an ISRU oxygen synthesizer, Perseverance is a versatile scientific platform [7], [8]. Its six-wheeled chassis features a combination of front and rear wheels with steering capability and a rocker-boogie suspension system, enabling speeds of up to 0.042 m/s.

ZHURONG (CNSA): Landing on Mars in May 2021 [9], the Zhurong rover is dedicated to uncovering evidence of frozen water on Mars while also characterizing the planet's surface and atmosphere composition. With dimensions of 1.23m x 0.83m x 540m (length, width, height) and a weight of 240 kg [10], Zhurong relies on an active boogie-rocker suspension system, equipping all six wheels with active steering for versatile locomotion modes [11]. It achieves a maximum speed of 0.055 m/s [10].

ROSALIND FRANKLIN (ESA): Positioned as Europe's pioneering Mars rover, the Rosalind Franklin rover, although fully developed, awaits its launch [12]. Its mission centers on determining the existence of past life on Mars and involves an array of instruments, including neutron and infrared spectrometers, an organic molecule analyzer, a Raman spectrometer, and a drill capable of reaching depths of 2 meters [13]. Rosalind Franklin measures 2.5m in width, 2m in height, and weighs 310 kg [14]. Its locomotion relies on six-wheel modules with steering capabilities and a "knee" joint for wheel-walking, integrated into a 3-boogie passive suspension system [15]. The rover is anticipated to operate at an average speed of $7.89*10^4$ m/s [14].

## 3. EMRS High-level architecture

A Theory section should extend, not repeat, the background to the article already dealt with in the Introduction and lay the foundation for further work. In contrast, a Calculation section represents a practical development from a theoretical basis.

### 3.1 Locomotion system

Within the locomotion system, four key elements have been delineated: Mobility, Steering, Suspension, and Chassis. Mobility encompasses the propulsion system, exemplified by elements such as wheels and wheel actuators. In this study, we have excluded discrete solutions (e.g., leg-based locomotion) and hybrid approaches (e.g., legs equipped with wheels) in favour of a continuous mobility solution, specifically prioritizing wheels. This choice is driven by the well-established maturity of wheel-based systems and their favourable mass-to-wheel performance ratio.

#### 3.1.1 Wheels

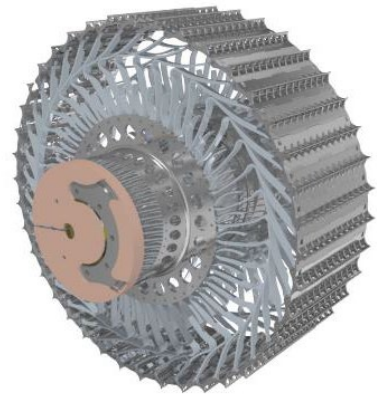

Figure 1 EMRS wheels developed by HTR.

Wheels are a critical component for the rover considering the harsh conditions of the lunar surface and regolith. They must provide traction on low-compacted sand, be able to overcome small obstacles, and withstand the erosion throughout the mission. The wheels developed for this project stand out for their high flexibility, which could be adjustable in the flight model, and the use of materials able to withstand the range of lunar surface temperatures without having their mechanical properties substantially diminished.

### 3.1.2 Steering

The term "steering" pertains to the mechanism facilitating the rover's ability to turn. Our investigation has identified two potential wheel independent steering options: "on-top steering" and "on-side steering" (right). Although on-top steering is usually preferred as it mitigates skidding during rotation and reduces the required rotational envelope; we have opted for on-side steering due to its compatibility with the upper payload bay volume requirements for diverse missions and to reach a more compact stowed configuration. The caveats of this configuration are dealt by considering this geometry in the kinematics.

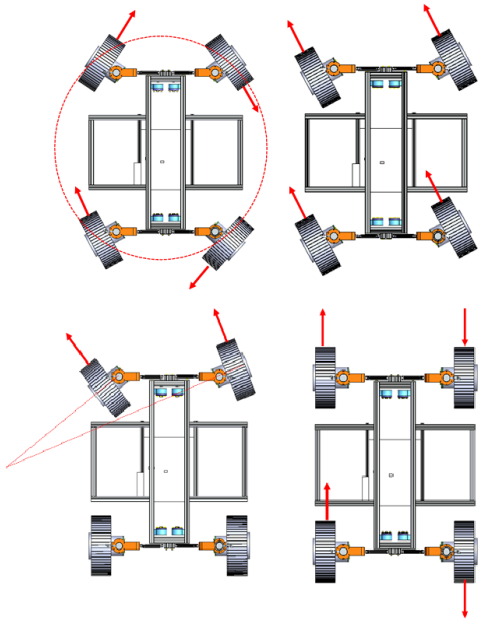

Figure 2 Steering modes.

The steering units allow each wheel to change their orientation independently up to +-90 deg, allowing the rover to have different locomotion methods.

Figure 3: Different locomotion modes enabled through independent wheel steering: Point turn (top left), Crab (top right), Ackerman (bottom left), and Skid Steering (bottom right).

### 3.1.3 Suspension

Suspension is designed to distribute loads effectively between the wheels during traversals, ensuring traction in challenging terrains, such as navigating over rocky surfaces. This aspect holds particular significance on the Moon, where the combination of low gravity and variable terrain compactness poses stability risks during rover movement. Our study has explored passive and active suspension solutions documented in the existing literature. We ultimately adopt a hybrid approach, incorporating a baseline passive solution featuring a parallelogram articulated fork with pre-loaded elements. Additionally, we offer the flexibility to implement an active suspension system atop the passive suspension, with minimal impact on the overall rover platform. This option remains reserved for missions with demanding scenarios, such as ISRU missions that may involve transporting up to 300 kg of regolith at certain points, potentially affecting rover stability. The active suspension allows for the adjustment of the Center of Gravity (CoG) when the rover is fully loaded, ensuring stability. Moreover, the active suspension enables a non-standard traverse mode known as "paddling," which proves particularly valuable for extricating the rover from soft terrains where it could become stuck, as well as allowing the rover to obtain a fully stowed configuration.

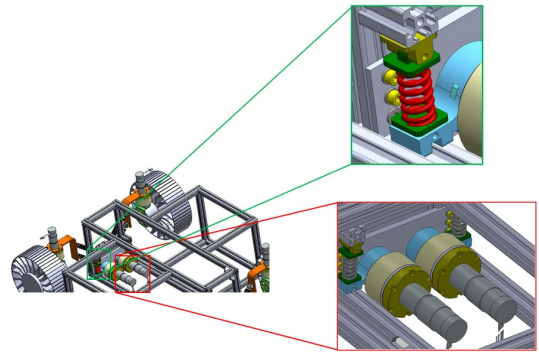

Figure 4 Passive and active suspension.

## 3.2 Chassis and structure

The chassis solution proposed is a modular structure made of CFRP. This modularity resides in the versatility of the locomotion system, and on the compartmentalization of different rover volumes to accommodate the different payloads and subsystems as its shown on .Figure.

The main bay is placed in the bottom centre of the rover, common to all the missions, will distribute the loads of the rover to the locomotion system and accommodate all the common avionics, thermal and power systems. To each side of this bay, lateral payload modules will take advantage of the space left in between the locomotion working envelopes to carry payloads, prioritizing those that need direct access to the lunar surface, such drills/excavators, or GPRs. Lastly, on top of the main receptacle, another payload volume is available to include instruments such as robotic arms, camera masts or antennas to name a few. These two configurable payload volumes change between the different missions allowing to reuse the rest of the rover elements.

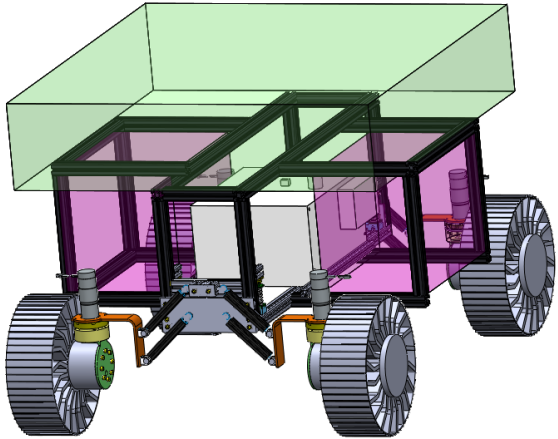

Figure 5 Preliminary design of proposed solution with the modular side payload volumes (pink) and top modular payload volume (green).

### 3.3 Payloads

Although the main avionics and many other versatile payloads may be common amongst the different EMRS missions, the specificity of each requires of specialised tools or instruments particular to the tasks to be performed.

The availability of the aforementioned payload volumes, adaptability of the locomotion system to different loads, and the overall modular design philosophy of the EMRS rover, allows the reusability of a single barebone rover design common to all missions. Each one of this, will then include particularities summarized as the following:

1. PE: carries scientific instruments to have FoV of the lunar surface, and as close as possible to the ground. This will be possible on the lateral payload volumes.
2. ISRU: The most demanding in terms of loads, it will need to carry up to 300 kg of lunar regolith, plus an excavator to collect it. This will be possible thanks to the active suspension that will adapt its stiffness as the load increases.
3. ALO: The deployable antenna hubs raise the CoG of the whole rover, challenging stability in sopes or when small obstacles tilt the rover. The passive suspension, flexible wheels and the improved stability thanks to the placement of the steering axes off-centre from the wheels mitigate this problem.
4. LGEM: A full scientific suite of sensors and actuators take advantage of the placement of the payload volumes of the rover, allowing for those which need direct access and proximity to the lunar surface to be placed in the lateral bays, and those that need to raise above the surface such as panoramic stereo cameras, or robotic arm carrying sampling and measuring instruments to have a free workspace to move around the surroundings of the rover.

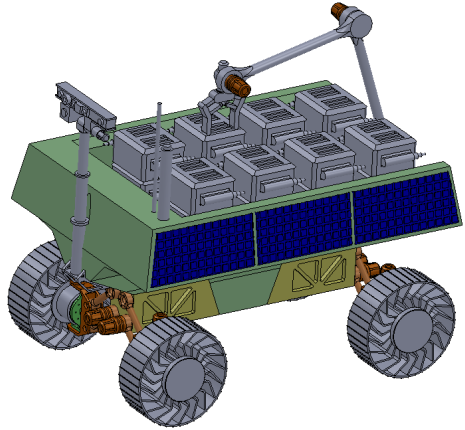

Figure 6 Example of EMRS rover system with ALO payload.

### 3.4 GNC Software and Autonomy

This software is implemented in order to control the degrees of freedom of the rover to be able to move with the different locomotion modes. The flow chart of the system is described in Figure X.

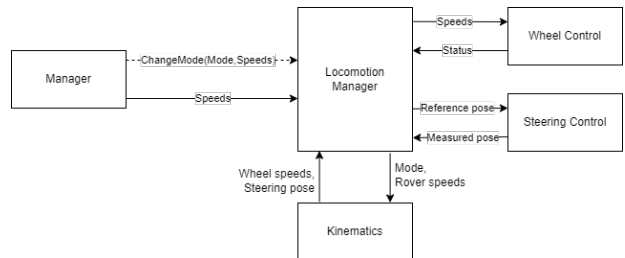

Figure 7 Block diagram of the locomotion software implemented in the avionics.

The different modules have the following characteristics:
- **Manager:** Interface module to receive the different commands available on the system. Separated in speed and change kinematic mode command. These inputs are prepared to be ordered by a joystick controller.
- **Locomotion Manager:** Responsible of synchronisation of the flow system. Getting the desired setpoints from the kinematics and generating the trajectories commands for steering and wheel control modules. Executing the changes of locomotion modes and supervising the well behaviour of the whole system.
- **Kinematics:** Designed to convert the vehicle speed commands into motor commands and the inverse conversion to obtain a locomotion odometry. Several locomotion modalities can be performed thanks of the rover architecture. The different modes are Skid Steering, Point Turn, Ackermann Steering and Crab Steering.
- **Wheel Control:** Module to execute a velocity control loop for the wheels motors as well as a security analysis of their behaviour.
- **Steering Control:** Module to execute a position control loop for the steering motors as well as a security analysis of their behaviour.

On the other hand, autonomy stands as a pivotal attribute in both current and prospective robotic applications, spanning the domains of space exploration and terrestrial operations. The crux of autonomy resides in endowing systems with the inherent capacity to make decisions independently [16]. This facet becomes particularly paramount when dealing with multifaceted robotic systems tailored to execute a diverse array of missions, each with distinct objectives and subsystem requirements. The endeavour of designing a rover capable of adapting to a spectrum of missions necessitates the development of software with the ability to abstract from mission-specific constraints. In essence, this entails crafting a goal-driven autonomous system.

The crux of achieving this lies in furnishing the rover with dynamic re-planning capabilities, coupled with semantic segmentation functionalities for the nuanced detection and identification of varied obstacles and objects, characterized by divergent shapes, sizes, and weights. Furthermore, the rover boasts manipulation capabilities that enable it to undertake tasks pertaining to In-Situ Resource Utilization (ISRU), manipulate tools, and execute scientific sampling operations.

Central to this robotic prowess is the avionics box, strategically nestled within the rover's structural framework. This housing unit harbours a pair of FPGA-based onboard computers, which shoulder the profound responsibility of overseeing critical functions encompassing guidance, navigation, and vehicle control [17]. Simultaneously, they handle housekeeping tasks, manage onboard data processing, and orchestrate telemetry and telecommand systems, all the while orchestrating seamless communications with other subsystems. Notably, these onboard computers also assume a pivotal role in driving the autonomous software framework that endows the rover with the prowess to conduct opportunistic scientific investigations.

Noteworthy enhancements to the rover's capabilities include stereo vision for localization and mapping, as well as AI-powered scene segmentation, path planning, and trajectory control. Leveraging the multifaceted potential of Artificial Intelligence (AI) affords us the capability to optimize the identification and prioritization of lunar resources. In doing so, we not only enhance the efficiency of In-Situ Resource Utilization (ISRU) activities but also significantly mitigate risks and bolster safety during these operations. The symbiotic integration of AI technologies amplifies the rover's capacity to operate autonomously, unlocking new horizons in scientific exploration and resource utilization.

## 4. Test Campaign

The overarching objective of the planned planetary analogue test campaign for the EMRS terrestrial prototype rover is to demonstrate the functional capabilities of the proposed rover concept, focussing on the locomotion system, in a relevant Planetary analogue environment. The place selected to perform the test campaign was the indoor soil bed of the DLR Planetary Exploration Laboratory.

*4.1 Facilities*

In accordance with the project's schedule, the testing facilities employed for conducting the required assessments were the DLR PEL (Planetary Exploration Laboratory) located in Weßling, Germany, and the GMV SPoT (Surface Planetary Terrain) facility situated in Madrid, Spain.

The selection of these two facilities aligns with the project's overarching goal of ensuring precise and thorough testing to validate the performance and functionality of the systems under development. This strategic choice of testing sites reflects the project team's commitment to achieving the highest standards of technical rigor and precision throughout the project's execution.

*4.1.1 DLR PEL*

The test area is a 5.5m x 10m indoor soil box (Figure X), used for testing planetary locomotion systems and navigation algorithms. A portion of the soil box (the rear 3.5m) can be tilted up to an angle of up to 30° to simulate steep slopes as can be shown in Figure Y. Additionally, rock and step obstacles in various sizes and shapes are available for testing obstacle clearance and agility, as well as providing realistic obstacles for navigation tests.

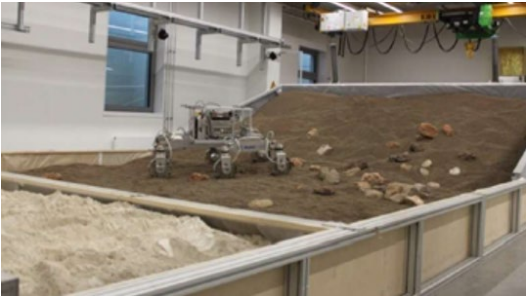

Figure 8 Indoor soil bed test facility

A variety of soil simulants are available, according to the needs of the test. The soil simulant has been selected as the 'basaltic soil' variant, in favour over the typical Eiffel lava material used by DLR. This consideration has been made to ensure the maximum representativeness of the simulant is achieved for the tests, given the understood soil characteristics of the lunar regolith expected. Details on the basaltic soil to be provisioned is given in [18].

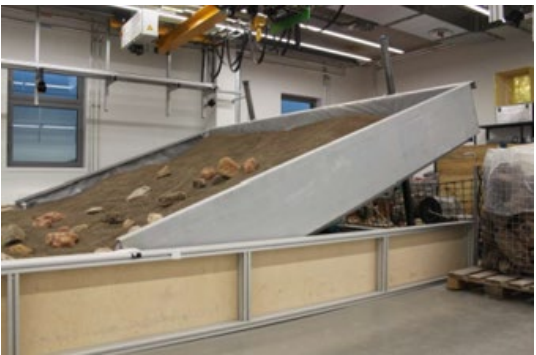

Figure 9 Soil bin with tilting portion

A pose tracking system is installed, consisting of 8 cameras, which can perform real-time tracking of the rover in 6 DoF at 60Hz, allowing for the ground truth rover position to be known at all times during execution of the test cases, up to an accuracy of approximately ~1mm and ~1° for position and orientation, respectively. Markers can be placed on the rover body or on landmark objects.

*4.1.2    GMV SPoT*

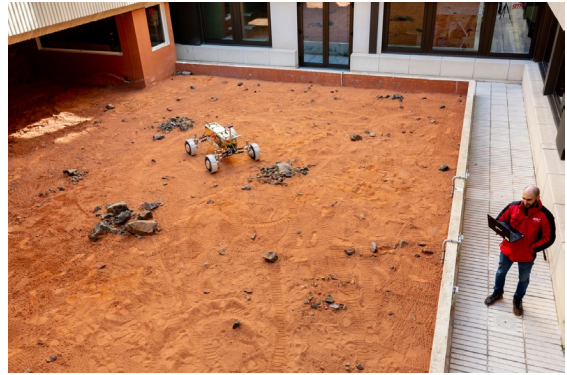

Figure 10 GMV SPoT

SPoT is a planet surface terrain facility which simulates a Martian landscape with similar soil and rocks. This facility is at GMV's head office, and it has a size of 180m$^2$, which provides a large testing area and an outdoor environment to test different robotic applications under natural lighting conditions from an annex serving as a space control centre.

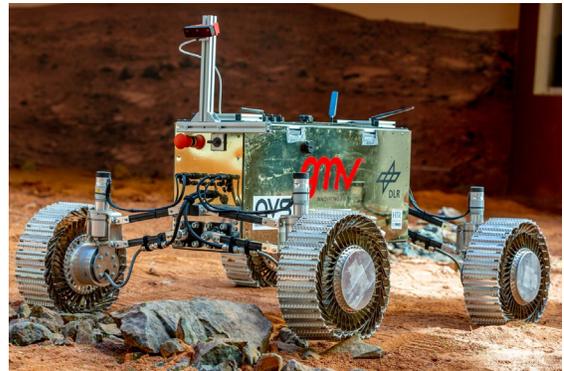

Figure 11 Traversability test at SPoT

The Table X summarize the parameters of the SPoT facility which should be considered for the prototype locomotion performance. The soil granularity and composition can affect at the wheel deformation and the skid in the differential drive mode.

Table 1. GMV SPoT soil characteristics.

| Parameter | Value |
|---|---|
| Size (m) | 15 x 12 |
| Height (m) | 0.12 - 0.2 |
| Specific weight (kg/m$^3$) | 1300 |
| Friction angle (kPa) | 8 - 12 |
| Deformation (deg) | 25 – 32 |
| Poisson ration (MPa) | 10 |
| Granulometry distribution (UNE-103-302-94) (mm) | 0.01 – 5 |
| Chemical composition of soil (%) | $SiO_2$  46<br>$Al_2O_3$ 27<br>$FeO_2$  7 |

*4.2 Test Methodology*

The testing activities were focussed on demonstrating the rover's locomotion system, where the terrestrial rover was commanded in a teleoperated manner to perform predetermined traverses in accordance with the scope of the defined test case.

The rover has to be able to accomplish a good performance on the following different manoeuvres:

- **Locomotion modes:** the rover has to perform the different locomotion modes explained in section 3.4.
- **Flat surface locomotion:** the vehicle has to perform a series of traverse segments at a variety of linear velocities with the soil bin completely flat.
- **Up-slope climbing:** the rover will drive up the slope of the soil bin raised to a defined angle.
- **Cross-slope locomotion:** the vehicle must perform a drive perpendicular to the slope direction of the soil bin raised to a defined angle.
- **Obstacle clearing**: the rover will traverse over different obstacles with different shapes and dimensions on a flat terrain.
- **Excavator**: attached with the excavator blade mock-up the rover has to perform a straight line at a representative linear velocity to demonstrate the level of traction available.

## 5. Field Test Campaign Results

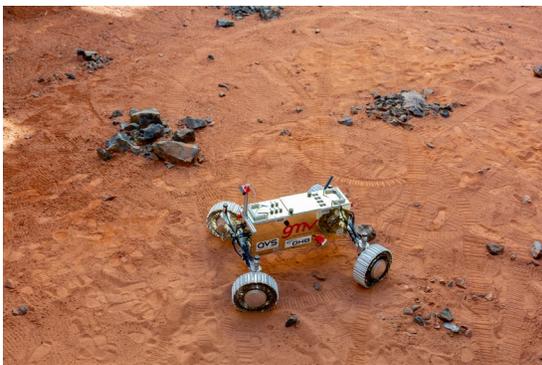

Figure 12 EMRS performing Ackermann steering in SPoT.

The test campaign performed in both DLR and GMV has served to validate the developed concept, with very promising results and performance above the expectations. The rover was able to perform all the maneuverer explained in the previous section.

All locomotion modes were able to execute and switch between them with any difficulties (Figure 13). The performance of all the locomotion modes behaves properly with the regolith avoiding standstills and long distances without the system getting stressed. The tests done overpassing obstacles were perform with irregular shapes and a height equal as the radius of the wheels.

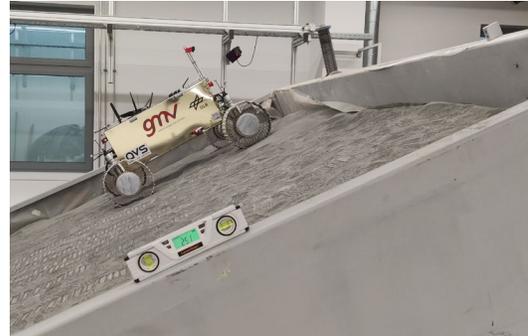

Figure 13 EMRS performing up-slope climbing in PEL with a tilt angle of 25 degrees.

The results obtained on the traverse done in the slope area in PEL shows that the rover can perform different maneuverer (up-slope climbing, cross-slope, point turns, Ackermann steering drive) with different angles. The tests were executed in slopes of 5, 10, 15, 20 and 25 degrees all performing properly. The wheels behave correctly avoiding slipping on up-slopes and cross-slop at the different slopes. Only point turning at a inclination over 20 degrees carry out a significant slipping.

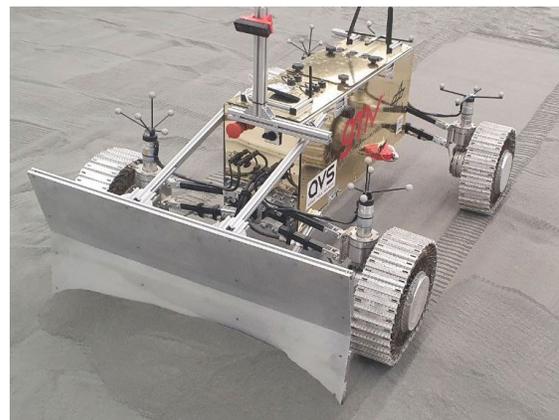

Figure 14 EMRS performing excavation tests in PEL

The excavation testing was accomplished as well with good results. The rover was able to perform several tests driving in a straight line in both directions. The rover didn't get stuck, and the system didn't get stress outside safety limits on temperature and electrical power performing always a smooth drive.

## 6. Conclusions

This paper introduces the versatile modular rover concept that has been developed as a part of the EMRS project. In the course of this project, we have not only conceptualized the rover but have also undertaken the design of critical systems encompassing power management, thermal control, insulation, and dust protection. It's worth noting that these systems will serve as the focal point of forthcoming investigations and refinements.

A series of rigorous obstacle and excavation tests have been conducted, shedding light on the remarkable capabilities of the EMRS system configuration. These tests not only demonstrate the rover's capability to navigate lunar terrain safely but also underscore its proficiency in excavating lunar regolith. The latter is of paramount importance in laying the foundation for sustainable human habitation on the Moon, as it enables essential resource extraction necessary for sustaining life and further scientific exploration.

The modular nature of our prototype design offers a unique advantage by affording us the opportunity to evaluate the rover's locomotion and software in conjunction with various scientific payloads. These payloads encompass a diverse range of instruments, including neutron spectrometers, drills, and a variety of cameras. This adaptability allows us to explore and fine-tune the rover's performance across a spectrum of scientific objectives, thereby maximizing its utility in future lunar missions.

**Acknowledgements**

The EMRS solution has been developed during European Moon Rover System Pre-Phase A project fully funded by ESA under grant agreement No. 4000137474/22/NL/GLC.